\documentclass{article}
\usepackage{PRIMEarxiv}
\usepackage{etoolbox}
\usepackage{fancyhdr}
\usepackage[utf8]{inputenc} 
\usepackage[T1]{fontenc}    
\usepackage{hyperref}       
\usepackage{url}            
\usepackage{booktabs}       
\usepackage{amsfonts}       
\usepackage{nicefrac}       
\usepackage{microtype}      
\usepackage{amsmath,amssymb,amsfonts}%
\usepackage{lipsum}
\usepackage{subcaption}
\usepackage{graphicx}       
\graphicspath{{media/}}     
\usepackage[table]{xcolor}
\usepackage{longtable}
\usepackage{floatrow}
\usepackage{caption}
\usepackage{xcolor}
\usepackage{upquote}
\usepackage{makecell}
\usepackage[export]{adjustbox}

\title{AI Model Passport: Data and System Traceability Framework for Transparent AI in Health}

\author{\hspace{1mm}Varvara Kalokyri \\
  Institute of Computer Science\\
  Foundation for Research and Technology Hellas\\
  Heraklion, Greece \\
  \texttt{vkalokyri@ics.forth.gr} \\
\And
  Nikolaos S. Tachos\\
  Biomedical Research Institute \\
  Foundation for Research and Technology Hellas \\
  Ioannina, Greece \\
  \texttt{ntachos@bri.forth.gr} \\
  \And
  Charalampos N. Kalantzopoulos \\
  Biomedical Research Institute \\
  Foundation for Research and Technology Hellas \\
  Ioannina, Greece \\
  \texttt{xkalantzopoulos@gmail.com} \\
    \And
  Stelios Sfakianakis\\
  Institute of Computer Science \\
  Foundation for Research and Technology Hellas \\
  Heraklion, Greece \\
  \texttt{ssfak@ics.forth.gr} \\
  \And
 Haridimos Kondylakis \\
  Institute of Computer Science \\
  Foundation for Research and Technology Hellas \\
  Heraklion, Greece \\
  \texttt{kondylak@ics.forth.gr} \\
  \And
  Sara Colantonio \\
  Institute of Information Science and Technologies\\
  National Research Council\\
  Pisa, Italy \\
  \texttt{sara.colantonio@isti.cnr.it} \\
  \And
  Daniele Regge \\
  Radiology Unit \\
  Candiolo Cancer Institute, FPO-IRCCS \\
  Candiolo (TO), Italy \\
  \texttt{daniele.regge@ircc.it} \\
  \And
  Nikolaos Papanikolaou \\
  Computational Clinical Imaging Group\\
  Champalimaud Foundation \\
  Lisbon, Portugal \\
  \texttt{nickolas.papanikolaou@research.fchampalimaud.org} \\
  \And
  The ProCAncer-I consortium \\
  \And
  Konstantinos Marias \\
  Institute of Computer Science \\
   Foundation for Research and Technology Hellas  \\
  Heraklion, Greece \\
  \texttt{kmarias@ics.forth.gr} \\
  \And
  Dimitrios I. Fotiadis \\
  Biomedical Research Institute \\
  Foundation for Research and Technology Hellas  \\
  Ioannina, Greece \\
  \texttt{fotiadis@uoi.gr} \\
  \And
  Manolis Tsiknakis \\
  Institute of Computer Science \\
  Foundation for Research and Technology Hellas \\ 
  Heraklion, Greece \\
  \texttt{tsiknaki@ics.forth.gr} 
}

\begin{document}
\maketitle

\begin{abstract}
The increasing integration of Artificial Intelligence (AI) into health and biomedical systems necessitates robust frameworks for transparency, accountability, and ethical compliance. Existing frameworks often rely on human-readable, manual documentation which limits scalability, comparability, and machine interpretability across projects and platforms. They also fail to provide a unique, verifiable identity for AI models to ensure their provenance and authenticity across systems and use cases, limiting reproducibility and stakeholder trust. This paper introduces the concept of the \textit{AI Model Passport}, a structured and standardized documentation framework that acts as a digital identity and verification tool for AI models. It captures essential metadata to uniquely identify, verify, trace and monitor AI models across their lifecycle - from data acquisition and preprocessing to model design, development and deployment. In addition, an implementation of this framework is presented through \textit{AIPassport}, an MLOps tool developed within the ProCAncer-I EU project for medical imaging applications. AIPassport automates metadata collection, ensures proper versioning, decouples results from source scripts, and integrates with various development environments. Its effectiveness is showcased through a lesion segmentation use case using data from the ProCAncer-I dataset, illustrating how the AI Model Passport enhances transparency, reproducibility, and regulatory readiness while reducing manual effort. This approach aims to set a new standard for fostering trust and accountability in AI-driven healthcare solutions, aspiring to serve as the basis for developing transparent and regulation compliant AI systems across domains.
\end{abstract}


\section{Introduction}

Artificial Intelligence (AI) is transforming healthcare by enabling the development of advanced tools that support clinicians in disease diagnosis, complex case analysis, and optimized treatment planning. However, for an AI system to be adopted in healthcare, it must be reliable, clinically useful, and above all, safe for patients \cite{WHO2021AIHealth}. Trust from healthcare professionals, patients, and policy makers is essential, and one of the key factors in building this trust is transparency \cite{gille, caspers}—ensuring that AI models are clearly documented, traceable, and their outputs are interpretable and accountable.

Transparency in AI is emphasized by multiple regulatory and scientific bodies, including the High-Level Expert Group (HLEG) on Artificial Intelligence \cite{eu2019ethics} and the European AI Act \cite{eu2021aiact}, which highlight the need for traceability, explainability, and clear disclosure of an AI system’s limitations. Additionally, the FUTURE-AI guiding principles \cite{futureai} emphasize that transparency must be built into AI development through structured documentation and monitoring of the entire AI lifecycle. This includes tracking data origins, processing steps, model training, and evaluation processes to ensure the system can be audited and validated. 

AI systems across healthcare rely on AI models trained on large-scale, often heterogeneous datasets. Variations in data sources, collection protocols, and processing methods can significantly impact the performance, reproducibility, and generalization capacity of these models, making comprehensive documentation essential. Maintaining data and model provenance is particularly important to identify and mitigate potential risks, such as the introduction of bias, which may arise from inherent imbalances in the data or from preprocessing decisions, including transformations and filtering applied by data scientists. Achieving transparency in such systems requires detailed \textit{data traceability}, including information on data ownership, collection methodologies, applicable standards, and curation procedures. Similarly, \textit{AI model traceability} demands thorough documentation of the entire model development lifecycle—from training and validation to deployment and monitoring. Both forms of traceability are essential for supporting transparency, as they enable effective auditing, reproducibility, and regulatory compliance \cite{Cantallops, Kroll}.

Transparency in the development and deployment of AI systems has attracted significant attention from regulatory bodies, the research community and industry stakeholders. While initiatives like Datasheets for Datasets \cite{Gebru2021}, Model Cards \cite{Mitchell2019}, and Google Data Cards \cite{datacards} promote structured documentation, they typically rely on manual, human-readable templates completed after development, and lack automation, standardization, and machine interpretability, limiting scalability and comparability across projects and platforms. Technical tools such as MLflow \cite{zaharia2018}, ML Metadata (MLMD) \cite{MLMD2019}, and OpenLineage \cite{OpenLineage2021} support metadata tracking during development but focus primarily on experiment management rather than providing end-to-end traceability. Although recent frameworks like ML Commons Croissant \cite{croissant} advance standardized metadata exchange for datasets, they do not address model provenance or identity verification. As a result, the lack of integrated solutions makes it difficult to reliably trace AI models' development histories, assess reproducibility, and verify authenticity across systems.
Enterprise solutions such as IBM FactSheets \cite{Arnold2019}, Azure Machine Learning \cite{AzureML2023}, and Google Vertex AI \cite{VertexAI2023} provide more comprehensive governance and monitoring capabilities, yet they rely on proprietary standards and fall short in delivering open, verifiable model identities that ensure provenance, authenticity, and compliance across diverse environments. This gap makes it difficult to verify a model’s development journey, assess reproducibility, and trace back the steps in its creation process.

To overcome current challenges in transparency and traceability, we propose the AI Model Passport—a standardized, machine-interpretable framework that automates the collection of essential metadata and monitors the entire AI model development lifecycle. Its aim is to bridge the gap between human-centric documentation tools and technical MLOps solutions and provide an integrated solution which focuses on recording critical information needed to uniquely identify, verify, and reproduce AI development processes. The AI Model Passport automatically tracks key metadata related to dataset creation and curation, model training and validation, and the stakeholders involved, ensuring full traceability and accountability throughout the workflow. By leveraging ontological models and integrating with open MLOps frameworks, it standardizes descriptions of datasets and AI workflows, enhancing interoperability, discoverability, and reusability of AI model assets. At the core of this framework is a unique, verifiable digital identity for AI models, enabling their verification across deployments. This identity-centric approach supports regulatory compliance, reproducibility, and fosters trust by ensuring models can be reliably identified, audited, and validated throughout their operational lifecycle. Furthermore, the framework is designed to align with the latest European regulatory initiatives and data governance strategies, including the Common European Data Spaces \cite{dataspaces}, the Open European Data Portal\cite{dataportal}, and demands of the European Health Data Space (EHDS) Regulation \cite{ehds_regulation}. These efforts emphasize structured, machine-readable documentation to promote trust, secure data sharing, and interoperability in digital ecosystems. In this context, the AI Model Passport also closely mirrors the principles behind the European Union’s Digital Product Passport  \cite{dpp_2024}, providing a standardized, verifiable mechanism to record and share critical lifecycle information—ensuring that AI models, like physical products, can be transparently traced, identified, and trusted across their usage environments.


Finally, we present and deliver \textit{AIPassport}, an open source, end-to-end MLOps pythonic tool, that implements the AI Model Passport framework and aspires to serve as the basis for developing transparent and regulation compliant AI systems across domains. Currently, deployed across several institutions, the AIPassport has supported the development of multiple clinical AI models within the ProCAncer-I European project. The ProCAncer-I project \cite{procancer}, an EU-funded initiative, integrates the largest database of anonymized multiparametric MR images related to prostate cancer (PCa), for developing robust PCa AI models and supporting precision care through the prostate cancer’s continuum. To showcase the efficacy of our implemenation, we demonstrate the AIPassport's practical use in developing a lesion segmentation AI model, leveraging data from over 14,300 patients collected through ProCAncer-I.

\section{Related Work}

Ensuring transparency, accountability, and ethical compliance in AI systems, particularly within healthcare, involves addressing several interconnected yet distinct areas: provenance metadata modeling, provenance tracking solutions, and identity management across the full AI lifecycle from data collection to model in production. Numerous frameworks, standards, tools, and ontological models have been developed, each contributing uniquely to these aspects.

Data provenance frameworks document the lifecycle of datasets, including origins, collection methods, ownership, and preprocessing transformations. The widely adopted W3C PROV Data Model \cite{prov-dm} provides a standardized representation covering entities, activities, derivations, agents, bundles, and collections involved in data production and usage. The Open Provenance Model (OPM) \cite{Moreau2008} represents provenance using a causality-directed acyclic graph (DAG), clearly depicting artifacts, processes, agents, and their interrelationships. The Provenance Vocabulary \cite{Hartig2010} specifically targets Linked Data, defining Actor, Execution, and Artifact classes to represent data creation, usage, and responsibility explicitly. Tailoring these foundations to the ML domain, the ML-Schema\cite{MLS} and PROV-ML\cite{provlake} models align experimental processes of AI workflows with formal metadata descriptions. More granular frameworks like the MEX Vocabulary \cite{mex} (composed of mex-core, mex-algo and mex-perf) offer detailed representations of machine learning experiments, covering aspects such as algorithms, configurations, and hyperparameters trying to address issues of managing ML outcomes and sharing provenance information. Expos{\'e} \cite{expose} supports standardized, interoperable sharing of experimental results, designed to describe and reason about ML experiments. The Data Catalog Vocabulary (DCAT) \cite{dcat}, although originally designed for dataset cataloging on the web, has increasingly been used to model metadata associated with input datasets to ML processes. In domains dealing with sensitive information, the Data Privacy Vocabulary (DPV) \cite{dpv} and its extension DPV-PD \cite{dpv-pd} provide semantic terms to describe data categories, processing purposes, and privacy risks, supporting regulatory compliance by embedding privacy-related metadata directly into dataset representations. Similar to this, the ML Commons Croissant standard \cite{croissant} has emerged recently as a way to support structured metadata sharing for improved interoperability and data sharing across institutions.

A diverse range of documentation frameworks, tools and systems has also emerged to document and operationalize provenance tracking and management. Enterprise-level frameworks such as FactSheets \cite{Arnold2019} and Google's Model Cards \cite{Mitchell2019} offer structured documentation formats that describe model goals, performance metrics, ethical considerations, and usage intentions. Frameworks such as Datasheets for Datasets \cite{Gebru2021}, DataCards \cite{datacards}, and Dataset Nutrition Labels \cite{Holland2020} further support transparency by capturing motivation, dataset origins, collection procedures, use cases and known limitations. Provenance-tracking tools and technologies like OpenLineage \cite{OpenLineage2021}, and Google’s ML Metadata (MLMD) \cite{MLMD2019} automatically capture metadata related to data pipeline executions and transformation steps. In the model development and evaluation space, MLflow \cite{mlflow} provides end-to-end experiment tracking, including parameters, metrics, and artifacts.  Cloud-native platforms like Azure Machine Learning \cite{AzureML2023} and Google Vertex AI \cite{VertexAI2023} offer integrated tools for bias analysis, explainability, and lineage tracing. Regarding technologies that support data, model and script/source code versioning and identification, GIT \cite{git} and Data Version Control (DVC) \cite{dvc} are the most widely adopted technologies that enable reproducible versioning of code, datasets, and trained models, being able to embed traceability and verifiability directly into AI pipelines. Regarding model deployment and operationalization, technologies such as MINIO \cite{minio} support object storage with versioning while also being and remain open sourced.

Despite these extensive efforts, many of the existing tools and frameworks remain isolated, addressing specific phases of the AI lifecycle without seamless integration. Ontologies, software tools, and documentation protocols often operate independently, creating fragmentation in traceability and compliance workflows. The AI Model Passport framework we introduce in this work aims to unify and extend the two worlds of human-centric documentation tools and technical MLOps solutions by integrating semantic ontologies such as PROV, DCAT, DPV, MLS with practical tools like DVC, GIT, MLflow, MINIO for lifecycle verifiable traceability. By combining structured metadata modeling with automated tooling, the AI Model Passport offers a standardized, interoperable solution that supports automated metadata collection enabling full traceability and model provenance, but most importantly a verifiable digital identity for the datasets (can also be seen as a ``Data Passport'' for AI related input datasets similar to the AI Model Passport), the final produced AI models, as well as all the development stages and code implementations for transparent, traceable, reproducible AI system development and deployment.
\section{AI Model Passport Framework}

In this section, the AI Model Passport framework is being described leveraging the work in the field. First, the underlying metadata model is formalized by the use of ontological frameworks, divided by the different distinct AI model development lifecycle stages. Then, we present the AI Model Passport technical infrastructure and the AIPassport tool.

The AI Model Passport Framework has been established based on the aforementioned detailed analysis of data and ML model provenance schemas, and on existing traceability tools. To define the framework, we employed a divide-and-conquer approach, examining each phase of the AI development and deployment process separately, and defining the data model and necessary processes starting from data selection to in-production model monitoring. 

\subsection{AI Model Development Lifecycle in Health}

According to Burov et al.\cite{burkov2020machine} a general rule of thumb for an AI model development lifecycle consists of: 


    \textit{Problem formulation:} Problem formulation requires the motivation to build and train a model. One should identify the absence or potential optimization of a service within an industrial or a societal environment, then proceed to define how an AI system can be used for it. For example, in the case of the healthcare domain, it is vital to define ``use case clinical scenarios'', to denote how AI can  provide support for the patients and clinicians, on various topics. One must consider also the ethical, legal, and regulatory implications of the problem.

    \textit{Data collection and preparation:} The second stage involves collecting relevant data and preparing them for use in training the AI system. This includes data cleaning, normalization, and transformation to ensure that the data are of sufficient quality to be used for training.

   \textit{Model development/evaluation:} The model development stage involves selecting appropriate algorithms and developing models based on the prepared data. The models are trained and tested to evaluate their performance.

    \textit{Model deployment:} Once the model is developed and tested, it is deployed in a production environment. This involves integrating the AI system into the existing IT infrastructure and ensuring that it meets the desired performance metrics.

   \textit{Model monitoring and maintenance:} After deployment, the AI algorithm must be monitored and maintained to ensure that it continues to perform at a high level. This includes ongoing monitoring of the algorithm's performance and feedback from domain users and stakeholders, as well as updating the algorithm as needed to address any issues.

    \textit{Retraining and retirement:} As data and their acquisition, along with the underlying phenomena, usually evolve and change over time, the AI algorithm may need to be retrained or replaced. This involves collecting and preparing new data, re-training the algorithm, and re-validating it for regulatory compliance.

\begin{figure}[!h]
  \centering
  \includegraphics[width=1\textwidth]{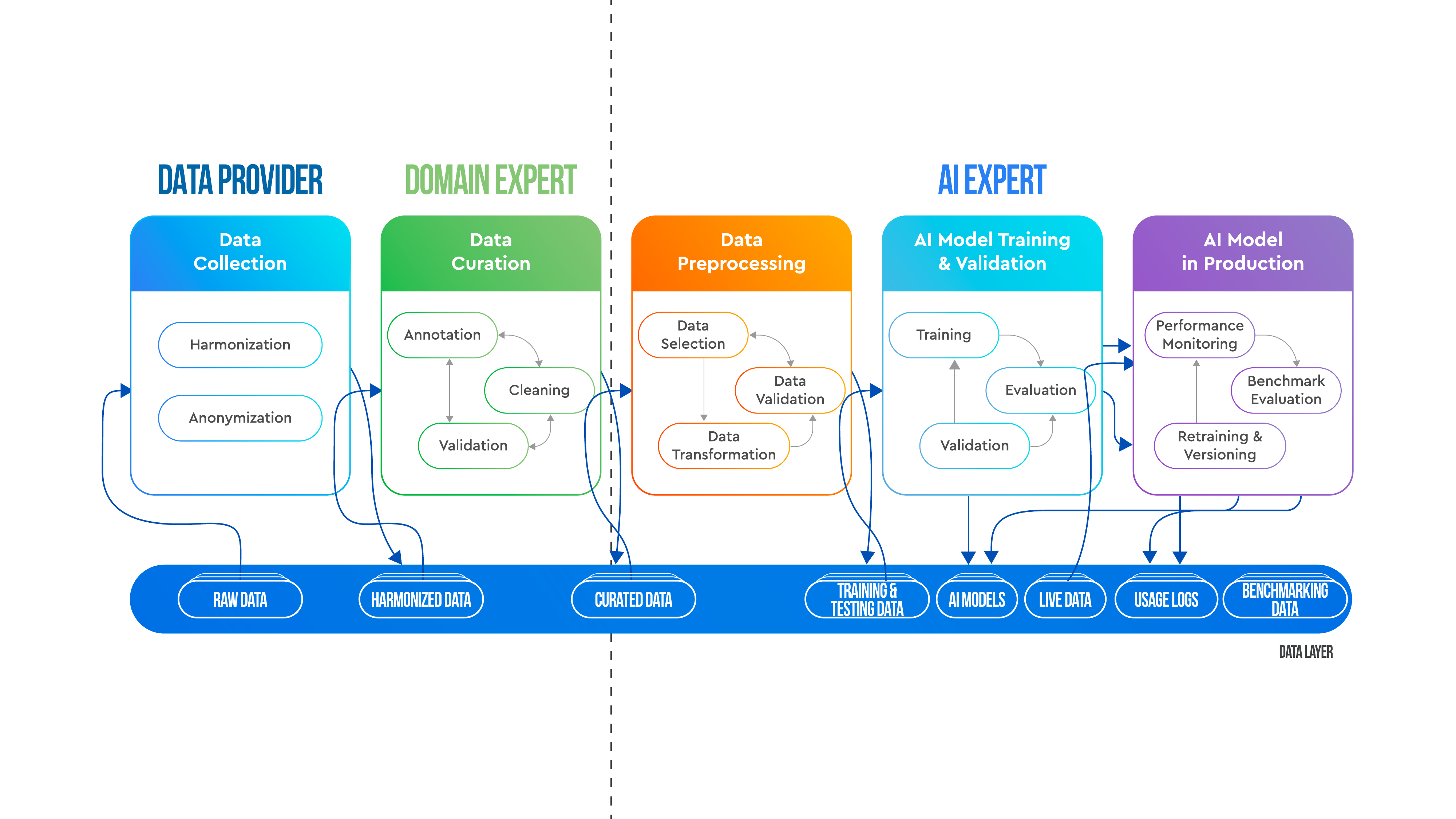}
  \caption{Generic AI development lifecycle in the health domain.}
  \label{fig:procancer_cycle}
\end{figure}

Different domains may require specific instantiations of these phases. In the health domain, the AI lifecycle is specialized to account for domain-specific data and workflows. In the following sections, we will describe in detail the various components and stages of this domain-specific AI lifecycle, as well as the data model used to describe the necessary provenance metadata at each stage, as shown in Figure \ref{fig:procancer_cycle}.

\textbf{Data Collection:}
In this stage, patient cases are gathered along with all the required clinical  data of various modalities, depending on the specific clinical scenario. Before data can be utilized, they  must be anonymized, harmonized, and standardized according to well-known clinical standards such as OMOP-CDM\cite{omop_cdm}, SNOMED-CT\cite{snomed_ct}, LOINC\cite{loinc}, FHIR\cite{fhir}, DICOM\cite{DICOM}, Radlex\cite{radlex} etc. Typically, data collection is conducted using a dedicated software tool, such as an electronic Case Report Form (eCRF), where data holders or data providers manage the information. 

\textbf{Data Curation:}
Usually, following the completion of the data collection process, a data curation phase is needed, which requires a significant amount of manual and highly specialized work, primarily conducted by domain experts. In the health domain, this includes data intensive workflows to clean, filter, and validate the data, or in specific sub-domains such as in radiology, to annotate medical images by radiology experts (e.g., performing segmentation tasks for specific organs, and/or possible cancerous lesions), applying motion correction on images that suffer from motion artifacts, or performing image co-registration tasks. Tracking this kind of data transformations is particularly important as this might affect the outcome of the AI models. For instance, the expertise and years of experience of the radiologists responsible for image segmentations may impact the quality of their outcomes, consequently influencing the performance of the trained algorithm. Therefore, it is imperative to thoroughly document every transformation process performed on the original raw collected data, including all the actors involved in the process.

\textbf{Data Preprocessing:}
There is a significant gap between the raw scientific data given and curated by clinical or domain experts and the useful data to be used as an input dataset to train ML models. To bridge this gap, a data preprocessing phase is essential. This phase aims first to restructure input datasets in a more organized and cleaner format, that is more comprehensible and easier to analyze before feeding the datasets into the AI model training process. In addition, a set of preprocessing tools and steps is usually being utilized by AI modeling experts at this stage after the datasets have been formed and accessed by the developers, addressing a set of different issues (e.g. bias field correction, denoising, image data harmonization etc.). For properly tracking all the data preparation/pre-processing procedures, a rich set of metadata that describe and track all the transformations being performed along with all the parameters, types of algorithms used in the processes, and the tools utilized is of paramount importance.

\textbf{AI Model Training and Validation:}
In this stage, AI model developers select the input training and validation datasets, choose training parameters that will be optimized in the training process, and select the “best” model depending on some evaluation metrics. Usually the entire pipeline that includes several preprocessing steps, feature selection, hyperparameter optimization is evaluated by the AI developers, who decide to apply more than one pipelines to compare, any may rerun them later with new data. It is important to mention that the AI Model Passport framework should be able to track metadata that refer to the final selected model, and not all the metadata from all the training runs, as this will create a big overhead in the metadata stored.


\textbf{AI Model in Production:}
Within the MLOps processes, monitoring models in production is a pivotal element, and in this context, a dedicated module for ongoing performance monitoring of deployed AI models is necessary. This monitoring module has the responsibility of swiftly identifying any deviation in the production model's performance from the expected metrics established during training and validation. Furthermore, upon detection or estimation of a decline in model performance, the monitoring module must trigger an alert to initiate a thorough investigation into the underlying causes. If necessary, it will facilitate the retraining of the AI models to rectify the situation.

In the following section, we will move forward in presenting a formal representation of the provenance metadata model employed for each of the aforementioned stages and the processes involved in each of them based on state-of-the-art approaches.
\subsection{Provenance Metadata Model Representation}
The metadata model used to track the whole AI development process follows the conceptual design of the AI Model Passport components described in the previous section. A key challenge here is the trade-off between capturing as much information as possible which causes high overhead in the capture of the provenance information, whereas not capturing enough information might lead to problems in reproducibility and understanding the actions/decisions that have been taken till the final production of the AI model. Therefore, striking a balance between the two options is necessary.
To achieve this, for each of the AI Model Passport components of the domain-specific AI lifecycle we described, we are delivering a metadata model to document a set of necessary information, which is compliant with standards such as W3C DCAT, W3C PROV and W3C MLS. Our modeling approach has focused extensively on minimizing human effort as much as possible in the population of metadata values within the AI Model Passport metadata model.

\subsubsection{Data Collection} \label{data_collection}
Data plays a critical role in AI services. The quality of a dataset used to build a model will directly influence the outcomes it produces. Thus, it is important to track the lineage of the dataset, as described by the many generic data provenance models. However, apart from the generic metadata, domain specific metadata related to the health domain we are dealing with should also be captured.



\begin{figure}[!htb]
  \centering
  \includegraphics[width=1\textwidth]{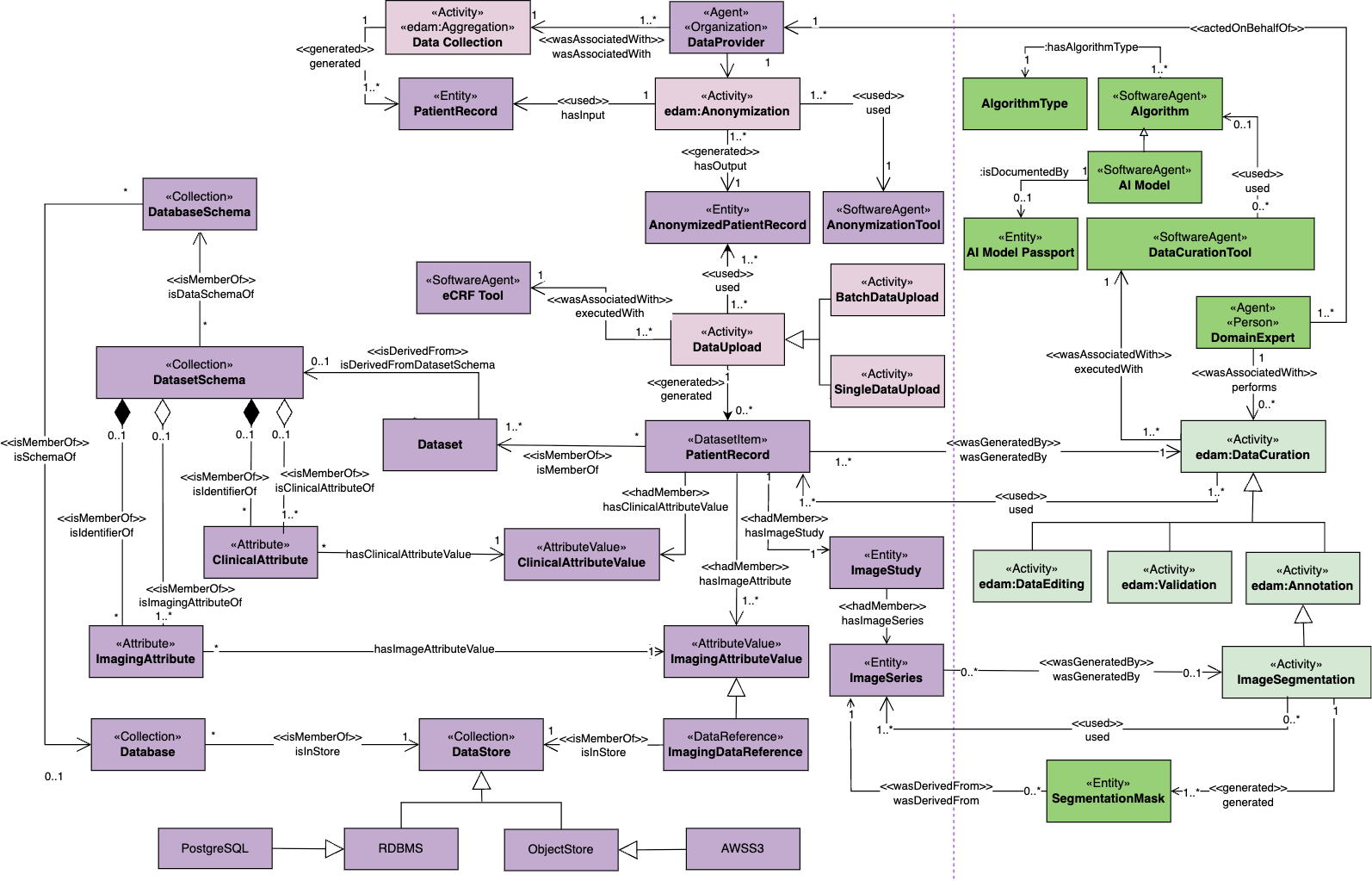}
  \caption{Class diagram representing the semantic structure of entities, activities, and agents involved in the AI data collection pipeline (on the left in purple) and data curation pipeline (on the right in green), extending PROV-O with domain-specific concepts.}
  \label{fig:data_representation}
\end{figure}

The conceptual model of the data collection component aligned with W3C PROV-O and PROV-ML schemas is shown on the left purple colored part of Figure \ref{fig:data_representation}. At its core, the model describes a \textit{PatientRecord}, which serves as a collection of clinical and imaging observations associated with a unique individual. Each record may include multiple \textit{ClinicalAttributeValue} and \textit{ImagingAttributeValue} entities, which are linked to corresponding \textit{Attribute} definitions (e.g., age at diagnosis, modality etc.). These values are connected through domain-specific properties like \textit{hasClinicalAttributeValue} and \textit{hasImageAttributeValue}. The imaging component is hierarchically structured, reflecting DICOM semantics: each \textit{PatientRecord} may reference one or more \textit{ImageStudy} entities, each of which contains \textit{ImageSeries}. This structure supports accurate linkage to the imaging data used in downstream analytical workflows. The provenance of patient data is maintained using \textit{prov:Activity} subclasses such as \textit{DataCollection}, \textit{Anonymization}, and \textit{DataUpload}, each performed by associated \textit{prov:Agent} entities like \textit{DataProvider} organizations or \textit{SoftwareAgent} tools (e.g., eCRF and anonymization software). These activities use and generate entities according to PROV relations such as \textit{prov:used}, \textit{prov:wasGeneratedBy}, and \textit{prov:wasAssociatedWith}.

Although Figure \ref{fig:data_representation} depicts a class diagram of the data collection component, based on the properties defined in the PROV model, the following object and data properties are being captured: the agent creating each patient record (individual or organization (\textit{foaf:organization})), the date of creation/upload, if the output of the activity has been validated/revised by someone (\textit{prov:wasRevisionOf}), the software used for uploading the information and the version of it (i.e. eCRF tool name and version of it), and finally if the data record has been invalidated (\textit{prov:wasInvalidatedBy}). In addition, some properties related to the specific patient record are captured, such as the anonymization method of the data anonymization activity if available (e.g. whitelisting, blacklisting), what is the clinical protocol used to upload the data record (e.g. retrospective/prospective cohorts) and the Use Case(s) to which the patient record belongs to in case the data collection process is being conducted as part of a specific use case that needs to be implemented.

All the patient records can be grouped into a bigger dataset, whereas those groups can be mapped to specific use cases of interest. However, depending on the needs and the use case, different datasets could be created for different purposes based on the same underlying data.

It should be noted that the metadata model described in this section corresponds to the data collected directly from the clinical sites (clinical information and imaging metadata at patient/record level), and it does not correspond to the dataset metadata model that is the direct input to the AI model (i.e., aggregated-level metadata describing the whole dataset as a resource). This information is being defined at a later stage of the AI pipeline, described in section \ref{section:preparation}.


\subsubsection{Data Curation}  \label{data_curation}
The component that follows the Data Collection is the Data Curation, where domain experts perform a set of data curation processes to clean, filter, further process, and annotate the data collected for the intended use case. The provenance model for data curation is shown in the right green part of Figure \ref{fig:data_representation}. 

As in this part of the workflow, the data curation is performed by domain scientists, it is also important to track the expertise of the individuals curating the data, and the years of experience they have, as this in many cases influences the quality of the produced dataset, and therefore the quality of the models trained by using this dataset. 

More specifically, in this model, every data curation process performed to the data, for example an image annotation process is captured based on the Data Curation activity and the associated entities, tracking specifically, who did the curation, what was the input and the output of the curation, if it was validated by someone, if any software was used for the data curation (along with its version, framework etc.), if there were parameters used in the curation process, and if the curation was done manually or automatically. 


\subsubsection{Dataset Specification} \label{section:preparation}
After the data collection and data curation processes are completed, the data must be organized and prepared for use by AI experts. Before this can happen, a dataset administrator must create one or more datasets, each associated with a well-defined set of metadata. These datasets should be assigned a unique, resolvable identifier that unequivocally links back to the data. This ensures that datasets used in AI model development are clearly identifiable, traceable if needed, and reusable for benchmarking across different models and studies. 

To facilitate the sharing, identification, and analysis of health data, a standardized framework for describing datasets is essential. Various initiatives and regulations have aimed to enhance data sharing across platforms and domains, making it easier for researchers and AI developers to find and assess datasets that meet their needs. For example, at the European level, efforts have been made to create centralized portals for cross-border dataset discovery. In the health domain, regulatory frameworks such as the EHDS regulation \cite{ehds_regulation} also emphasize the importance of structured dataset catalogues, ensuring accessibility and interoperability in digital health ecosystems.

To achieve this, a common framework is needed to describe datasets in a standardized manner, ensuring consistency across data catalogues. The W3C Data Catalog Vocabulary (DCAT) \cite{dcat} standard provides a generic, cross-domain model for structuring dataset metadata. The DCAT-AP (Application Profile) \cite{dcat_ap} extends this standard by introducing additional constraints and controlled vocabularies, improving interoperability across different data portals within the European Union. However, since the health domain—particularly in clinical applications—requires domain-specific metadata, a more tailored approach is necessary. Towards this, the HealthDCAT-AP specification \cite{healthdcat_ap} is currently being finalized in the context of the EHDS, which is recommended to be used for health related dataset descriptions, which will achieve interoperability of all health-related dataset descriptions in Europe.

\begin{figure}[!htb]
  \centering
  \includegraphics[width=1\textwidth]{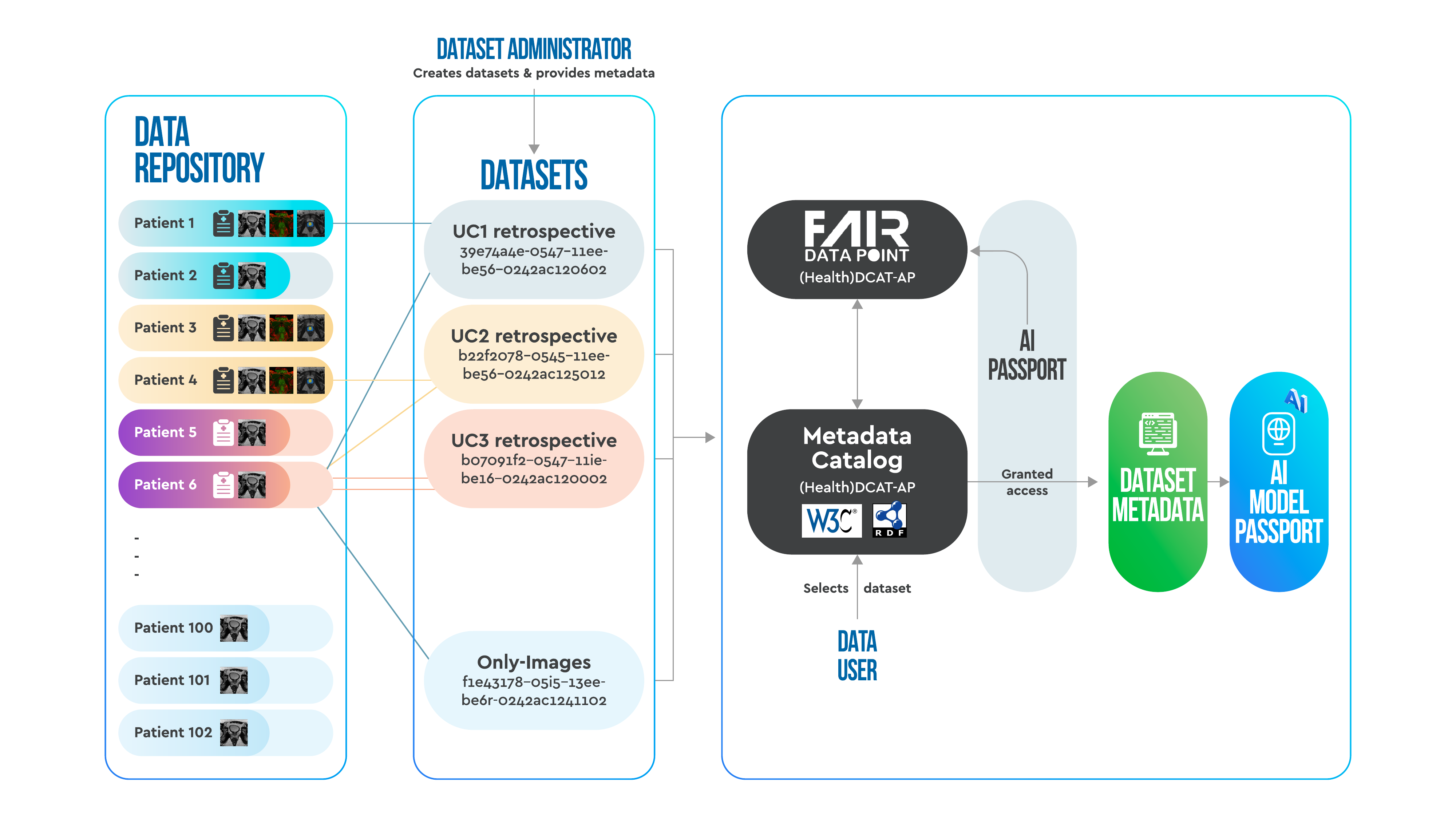}
  \caption{Workflow of dataset preparation and use in the AI Model Passport framework.}
  \label{fig:datasetPreparation}
\end{figure}

The AI Model Passport framework, designed to support compliance with the European emerging policies, proposes that all AI dataset descriptions should adhere to the HealthDCAT-AP specification and be exposed in a FAIR-compliant way through the use of Fair Data Points (FDPs)\cite{fdp}. A FAIR Data Point (FDP) is a service that publishes datasets in a standard way, that allows both machines and humans to find, access, and understand data with minimal effort. By leveraging the HealthDCAT-AP standard and the FDP services, all datasets to be used for AI model development are made available and, when permitted, downloadable to data users via the “Distribution” entity of the DCAT-AP specification. 

As such, the AI Model Passport framework has been designed to account for and exploit these standardized specifications as follows: An AI expert can browse a metadata catalog, select a dataset that fulfills their needs based on its description, and request access. Once access is granted, the AI Model Passport implementation (i.e. the AIPassport tool) can harvest the standardized metadata through the FDP, following the DCAT-AP specification, along with the corresponding dataset \textit{“Distribution”} (either by direct download or through controlled access mechanisms, such as a secure processing environment as envisioned by the EHDS regulation). This way, the input dataset metadata are automatically stored in the AI Model Passport documentation of the developed AI model. A representation of such workflow is depicted in Figure \ref{fig:datasetPreparation}.

 
 It is important to note that since DCAT is a generic model, the AI Model Passport framework is capable to process and understand metadata descriptions even if these are not from the health domain, paving the way for extending the framework for other domains.


\subsubsection{AI Model Training and Evaluation}

\begin{figure}[!htb]
  \centering
  \includegraphics[width=1\textwidth]{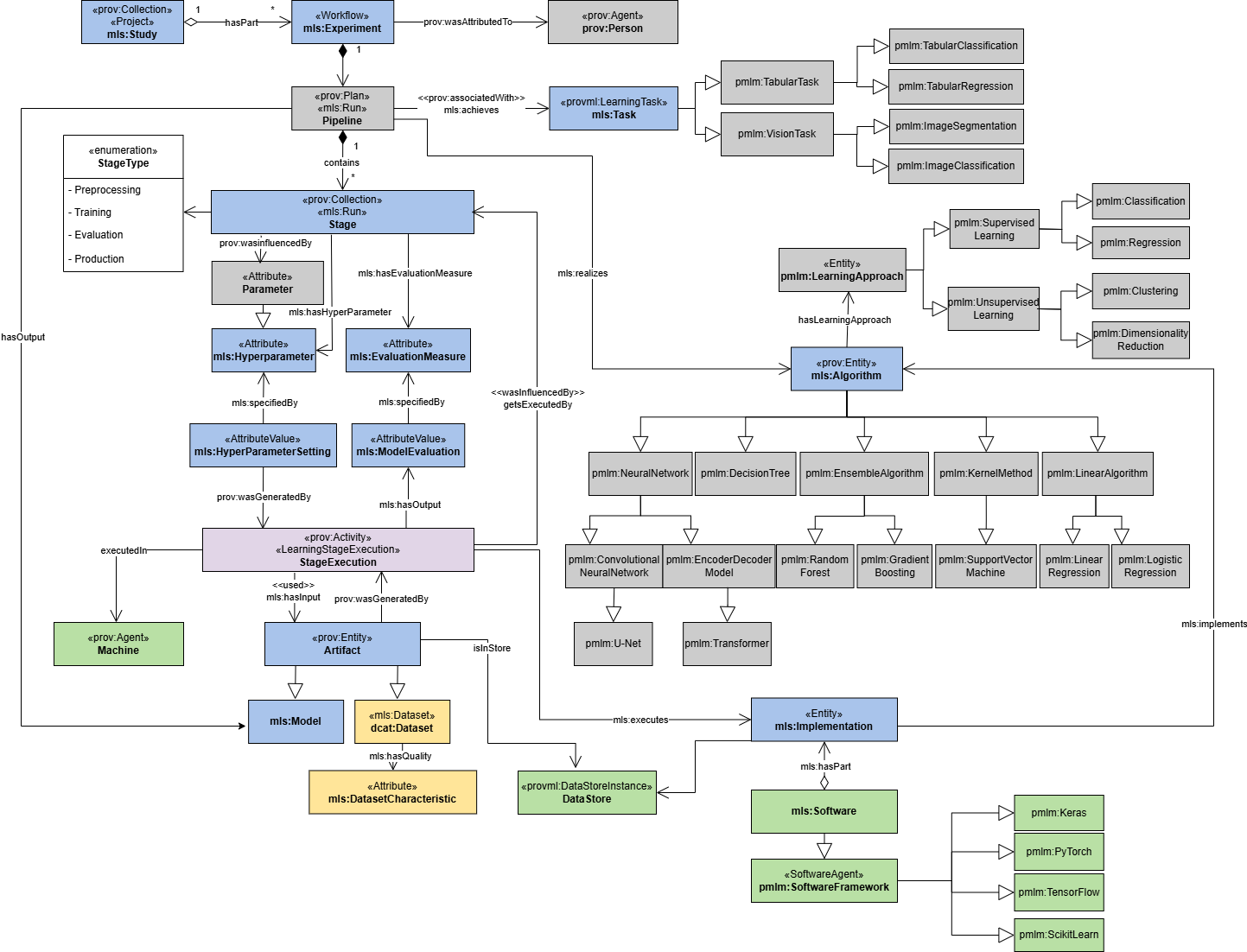}
  \caption{AI training and evaluation workflow provenance data representation}
  \label{fig:data_pretraining}
\end{figure}

Model training and evaluation are critical phases in the AI lifecycle, demanding rigorous documentation of computational steps, configurations, and resulting artifacts to support transparency, reproducibility, and trust. To formally describe this workflow, we employ a provenance-driven, workflow-centric model grounded in the W3C PROV and W3C MLS standards, as illustrated in Figure \ref{fig:data_pretraining}. In our representation, a \textit{Study} acts as a container for multiple {Experiment} instances, each of which comprises a single \textit{Pipeline}—a structured sequence of \textit{Stage} entities corresponding to key lifecycle phases such as \textit{preprocessing, training, evaluation, and production}. These stages are instantiated through \textit{StageExecution} activities, which capture the operational flow of each step.

The use of W3C PROV enables granular tracking of data transformations, computational activities, and responsible agents, thereby ensuring end-to-end traceability of the AI development process. Each \textit{StageExecution} records critical provenance information, including hyperparameter configurations, evaluation metrics, and the generation of datasets and models. To express machine learning-specific constructs, we integrate and capitalize the Provenance for Machine Learning Models (PMLM) ontology \cite{pmlm}, which aligns with and extends the PROV-O standard. Within this framework, \textit{pmlm:LearningTask} defines the nature of the problem being solved (e.g., \textit{pmlm:ImageSegmentation}), while \textit{pmlm:LearningApproach} distinguishes between paradigms such as supervised, unsupervised, and reinforcement learning. High-level algorithm families, such as \textit{pmlm:NeuralNetwork}, are captured under \textit{pmlm:Algorithm}, with specific model types—like \textit{pmlm:U-Net}—serving as concrete instantiations. These are implemented via \textit{mls:Implementation} entities, which encapsulate executable components along with metadata about software libraries (\textit{pmlm:SoftwareFramework}), such as \textit{TensorFlow} or \textit{PyTorch}, and associated runtime environments. Model evaluation is formally represented through the \textit{mls:ModelEvaluation} construct, linking evaluation results to specific metrics and contextualizing them within the overall experiment. 

\subsubsection{AI Model in production}
The final stage in the AI model lifecycle is its deployment in a production environment, where it is actively used for inference and decision-making. At this stage, the AI model is accompanied by a rich set of metadata, which largely mirror those captured during the AI model training and evaluation phase, with the key distinction that they now represent the final values corresponding to the best-performing trained model selected for deployment.

In the AI Model Passport framework, several metadata attributes are automatically recorded based on the system's internal tracking mechanisms, which will be described in the following section, including details such as hyperparameters, software dependencies, libraries used, and final evaluation metrics. Additionally, other metadata fields must be manually documented by the AI developer, capturing critical contextual information such as the specific learning problem addressed, the algorithm used, model descriptions, risk assessments, and potential ethical concerns. Most of these values are populated through standardized vocabularies to avoid ambiguity (such as PMLM, DPV).

Beyond static metadata, real-time monitoring metrics should also be collected to assess the model’s performance in production. These metrics include outlier detection and drift detection indicators, which track potential deviations in data distributions over time. By integrating these monitoring components, the AI Model Passport should maintain a comprehensive and dynamic record of model performance, ensuring end-to-end traceability, reliability, and adaptability to evolving clinical or operational conditions. However, in the current AI Passport implementation, monitoring metrics are not yet integrated but are planned as part of future development.
\subsection{AI Model Passport Implementation - AIPassport}
To address the challenges of transparency, traceability, and reproducibility in AI model development, the AI Model Passport framework has been implemented as an end-to-end MLOps framework - the \textit{AIPassport} \footnote{https://git.procancer-i.eu/ChKalantzopoulos/aipassport. Currently accessible only for the ProCAncer-I consortium}. The AIPassport is an end-to-end MLOps tool that automates the documentation of AI workflows. Implemented as a Python package, AIPassport captures and tracks all relevant metadata across the AI model lifecycle—from data preparation and preprocessing to training, evaluation, and model registration. It is designed to integrate seamlessly into a developer’s existing pipeline without imposing technical overhead, supporting flexibility across operating systems and frameworks (e.g., TensorFlow, PyTorch, Scikit-learn etc).

Built on top of open standards, AIPassport ensures semantic interoperability and compatibility with broader initiatives and promotes FAIR principles in machine learning by automatically versioning datasets and scripts, assigning unique identifiers, and recording provenance of all activities and artifacts. AIPassport also supports integration with established MLOps tools—namely, MLflow for experiment tracking and model registry, DVC (Data Version Control) for tracking changes in datasets and pipelines, Git for code versioning, and MINIO as a backend object store for models, logs, and data. The tool enables users to register a rich set of metadata, both technical and contextual. It documents dataset origins, preprocessing steps, model parameters, evaluation metrics, environment configurations, as well as subjective descriptions such as the model’s intended use or ethical considerations. While most of this information is gathered automatically, AIPassport provides a structured interface to guide developers in manually declaring information that cannot be extracted programmatically. 

\subsubsection{Architecture}

\begin{figure}[!htb]
  \centering
  \includegraphics[trim=0 250 0 0, clip, width=0.6\textwidth]{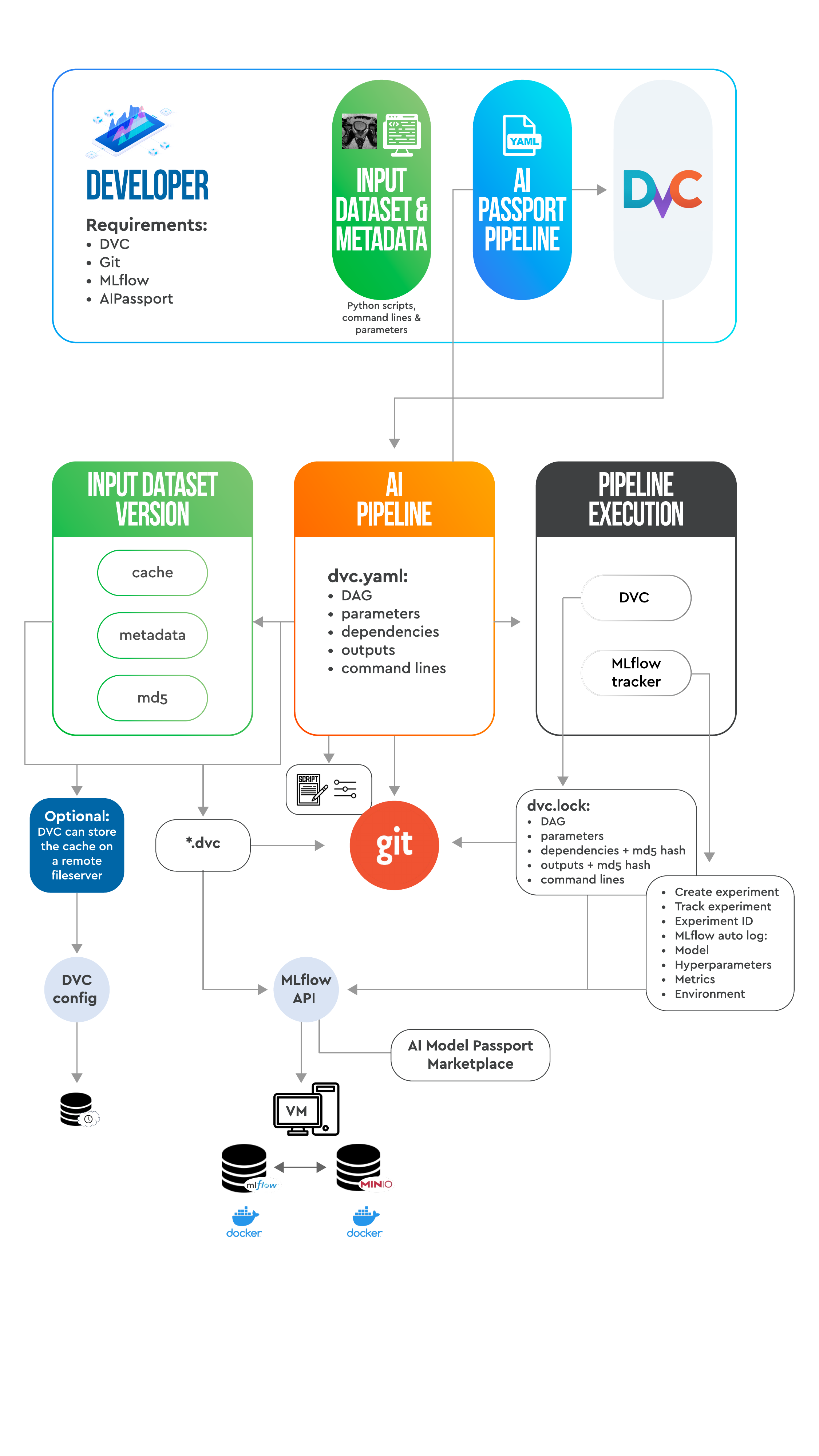}
  \caption{AIPassport infrastructure overview illustrating the integration of MLflow, DVC, Git, and MINIO to support versioning, experiment tracking, and metadata capture across the AI model development lifecycle.}
  \label{fig:infrastructure}
\end{figure}

The AIPassport framework follows a modular architecture illustrated in Figure~\ref{fig:infrastructure}, which is designed to operate in both on-premise and cloud-based environments. 

Its core infrastructure comprises three tightly coupled components that enable comprehensive model lifecycle management. The first component is the MLflow server, which provides services for experiment tracking and model registry. During the training phase, MLflow collects metadata about the dataset used, the hyperparameters configured, the performance metrics evaluated, and the final trained model artifact. The second component is the MINIO object storage service, which stores datasets, intermediate files, logs, and final model binaries. MLflow and MINIO communicate through an S3-compatible API, allowing scalable and secure artifact storage. The third component includes DVC and Git, which together manage dataset and code versioning, ensuring that each experiment’s dependencies, inputs, and outputs are explicitly recorded and verifiable.

AIPassport itself serves as the orchestrator that glues these components together. Upon initiation, it connects securely to the infrastructure using session tokens and sets up the pipeline execution environment. It then initializes Git and DVC repositories and creates a dedicated MLflow experiment space. During execution, AIPassport parses the user-defined pipeline stages and tracks each stage's configuration, input, and output. It captures parameters passed via command line or params.yaml (required by DVC), and logs the results, metrics, and artifacts into MLflow. Each pipeline stage—whether it involves preprocessing, training, evaluation, or model deployment—is treated as an independent, traceable step in the lifecycle.

AIPassport encourages the use of modular AI workflows, where each task with a distinct purpose is encapsulated in a dedicated Python script and described declaratively. Users define the pipeline structure—including inputs, outputs, parameters, and commands of each stage within a configuration file (.yaml). AIPassport consumes this definition to generate the DVC pipeline and to monitor its execution. As the pipeline progresses, intermediate datasets and artifacts are versioned and logged, facilitating reproducibility.

Finally, for model training or production stages, AIPassport leverages MLflow’s autologging functionality to automatically record model metrics, dependencies, and environmental conditions. For additional metadata that cannot be automatically inferred—such as model purpose, deployment constraints, or ethical considerations—a template is provided to guide users in documenting these manually.

In the next section, we present a concrete use case demonstrating how AIPassport has been applied in practice within the ProCAncer-I project, capturing the end-to-end development and evaluation of a prostate lesion segmentation model.

\section{Use Case in ProCAncer-I}

The ProCAncer-I project \cite{procancer} is an EU-funded project, currently comprised of more than 14,300 patient cases, and more than 9,5 million multiparametric MR images related to prostate cancer. The project covers various clinical scenarios, from diagnosing and characterizing prostate cancer to predicting treatment responses and potential side effects. To reply to such scenarios, clinical experts have detailed the requirements and provided specific guidelines for data collection, in order to ensure that medical images and their respective clinical metadata align with the unique needs of each clinical scenario, aiding in the development of effective AI models. 

In this section we will showcase the implementation of the AI Model Passport framework as a proof of concept for tracking the development of a lesion segmentation model.

\subsection{Data Collection}
Starting with the data collection component of the AI lifecycle, below we illustrate an example instantiation of the data collection stage from ProCAncer-I. 

\begin{figure}[!htb]
  \centering
  \includegraphics[width=1\textwidth]{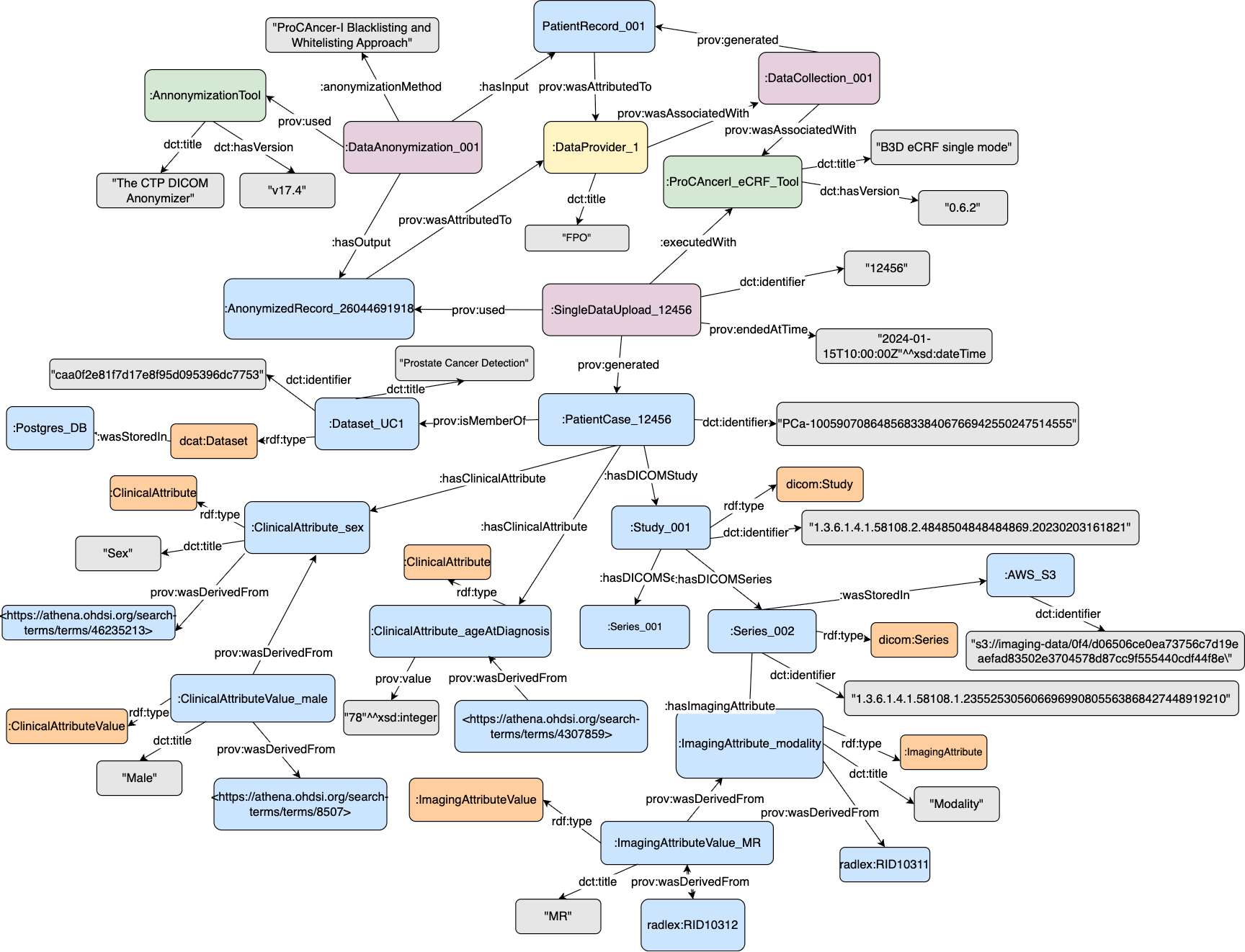}
  \caption{Instance-level provenance graph capturing the data collection lifecycle for a patient case. The graph models the transformation of a raw patient record into an anonymized dataset and ultimately into a structured patient case,  aligned with PROV-O domain specific clinical and imaging metadata.}
  \label{fig:data_collection_instance}
\end{figure}

Figure~\ref{fig:data_collection_instance} presents an instance-level provenance graph that traces the data collection process for a single patient case within an AI development workflow. It represents key steps such as a data collection activity associated with a clinical data provider, an anonymization activity made by a software agent, and finally the upload to a data repository. Each process is described as a prov:Activity (in purple), connected to its inputs and outputs using prov:used, prov:generated, and prov:wasPerformedBy. The resulting anonymized patient case is enriched with structured clinical and imaging attributes, each linked to standardized vocabularies such as OMOP and RadLex.

\begin{figure}[!htb]
  \centering
  \includegraphics[width=1\textwidth]{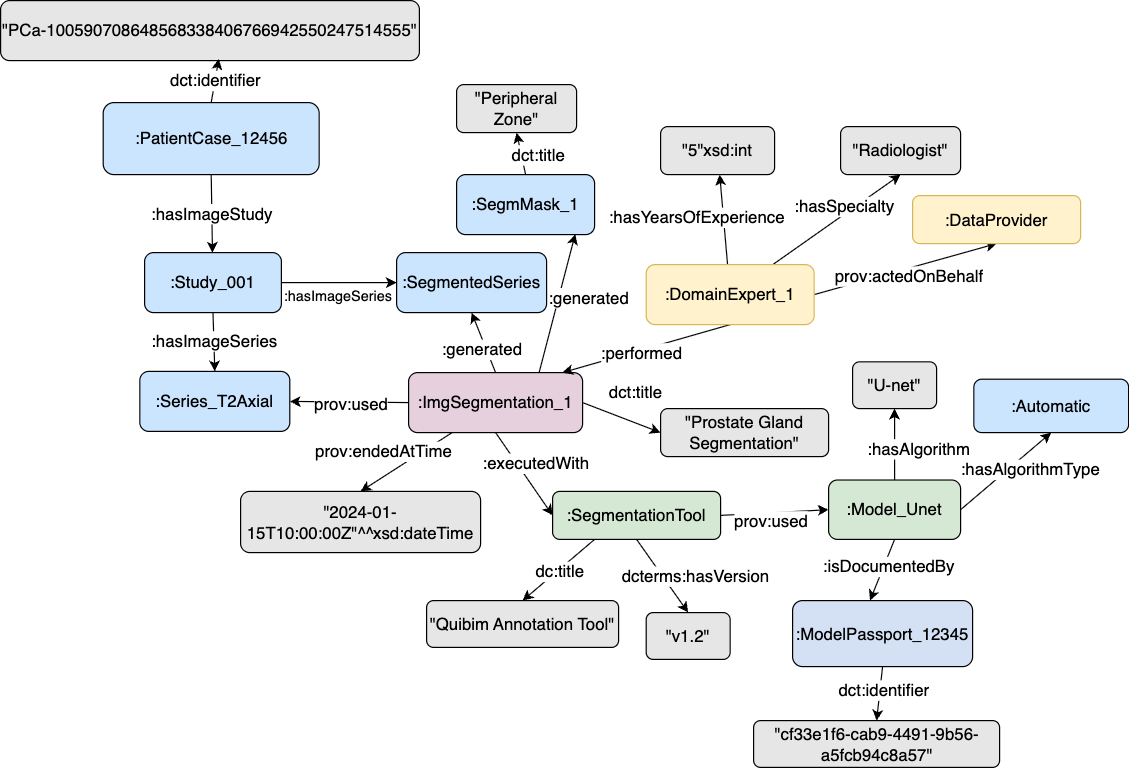}
  \caption{Instance graph of an image segmentation task, showing the input image series, segmentation activity, involved agents and tools, and the generated segmentation mask with its provenance.}
  \label{fig:data_curation_instance}
\end{figure}

\subsection{Data Curation}
Continuing with the data curation phase of the AI lifecycle, Figure~\ref{fig:data_curation_instance} illustrates an example instantiation of an image segmentation task as part of a data curation process. This instance-level provenance graph models the curation of a prostate MRI T2w series of patient through a segmentation activity. The figure captures the main steps including the input image series, the segmentation activity performed by a domain expert, and the use of a dedicated segmentation tool, which in this example uses an AI model to perform the task (automatic). The provenance graph shows how the segmentation mask is generated and explicitly linked to the original image series through prov:generated relationships. In this example, the AI model used in the process has also been documented through its own AI model passport (in case such a passport exists), supporting clear traceability.

\subsection{Dataset Specification}

Following the data collection and curation processes, the ProCAncer-I dataset administrator defines and describes the different datasets in order to be added to the ProCAncer-I metadata catalogue. 



\begin{figure}[!htb]
  \centering
  \includegraphics[width=\textwidth]{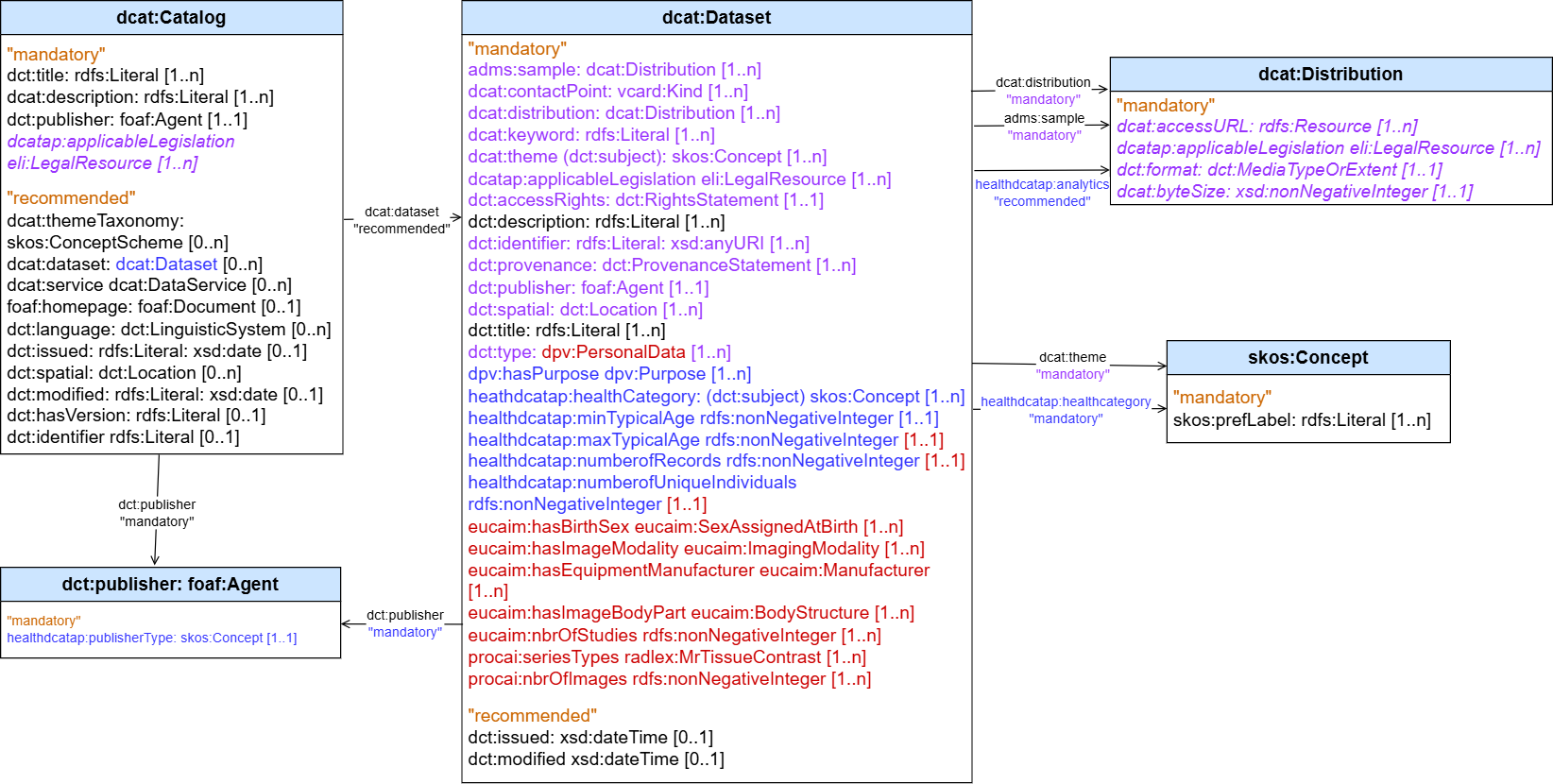}
  \caption{ProCAncer-I DCAT-AP extension for the prostate mpMR imaging datasets based on HealthDCAT-AP}
  \label{fig:dcat_ext}
\end{figure}

Figure~\ref{fig:dcat_ext} presents the core extension of the DCAT-AP specification as implemented in the context of the ProCAncer-I project. The diagram illustrates the hierarchical structure of the metadata catalog, including the Catalog, the associated Datasets, and their corresponding Distributions. To improve readability, optional fields have been omitted. The figure uses a color-coded scheme: properties already defined in the DCAT-AP specification are shown in black; existing properties with refined semantics are shown in purple italics; new attributes introduced by the HealthDCAT-AP extension appear in blue; and attributes specific to the cancer imaging domain—introduced by ProCAncer-I—are highlighted in red. For each newly introduced property, a controlled vocabulary has been defined, building upon those already proposed by the DCAT-AP standard.

In the current implementation of the AI Model Passport framework within the ProCAncer-I project, input dataset metadata are made compliant not only with the HealthDCAT-AP specification but also with the EUCAIM DCAT-AP extension. EUCAIM \cite{eucaim} is an EU co-funded initiative that aims to establish a pan-European federated infrastructure for cancer imaging. It provides a trustworthy platform for researchers, clinicians, and innovators to access diverse imaging data for benchmarking, testing, and deploying AI-based technologies. EUCAIM defines its own ontology\cite{eucaim_ontology} identified with the \textit{eucaim:} prefix and the namespace \textit{\url{https://cancerimage.eu/ontology/EUCAIM}} along with a specialized DCAT-AP extension tailored to the cancer imaging domain.

Furthermore, a FAIR data point of the ProCAncer-I catalogue has been implemented.
Through the ProCAncer-I FDP, the metadata are made available through http protocols in a machine readable format (.rdf turtle format). This way, all the ProCAncer-I datasets will gain visibility in the EHDS ecosystem, and be interoperable with the rest of the European health datasets. In addition, EUCAIM will automatically harvest the dataset metadata of the ProCAncer-I metadata catalogue and include them in its own public catalogue, for greater visibility. %

\subsection{Data pre-processing}

After the selection of the proper dataset by an AI developer for developing the lesion segmentation scenario and the proper access has been granted, the metadata that correspond to this dataset get downloaded through the platform along with the imaging and clinical data in a secure processing environment within the ProCAncer-I platform. 



After the dataset input metadata have been extracted and stored, the AI pipeline gets declared by the developer in a `params.yaml' file required by DVC. This is required in order for the DVC framework to create and store the pipeline as a directed acyclic graph (DAG), keep track of any changes of each of the pipeline stages along with the input and output data, and stage's parameters and configurations. 

\begin{figure}[!htb]
  \centering
  \includegraphics[width=1\textwidth]{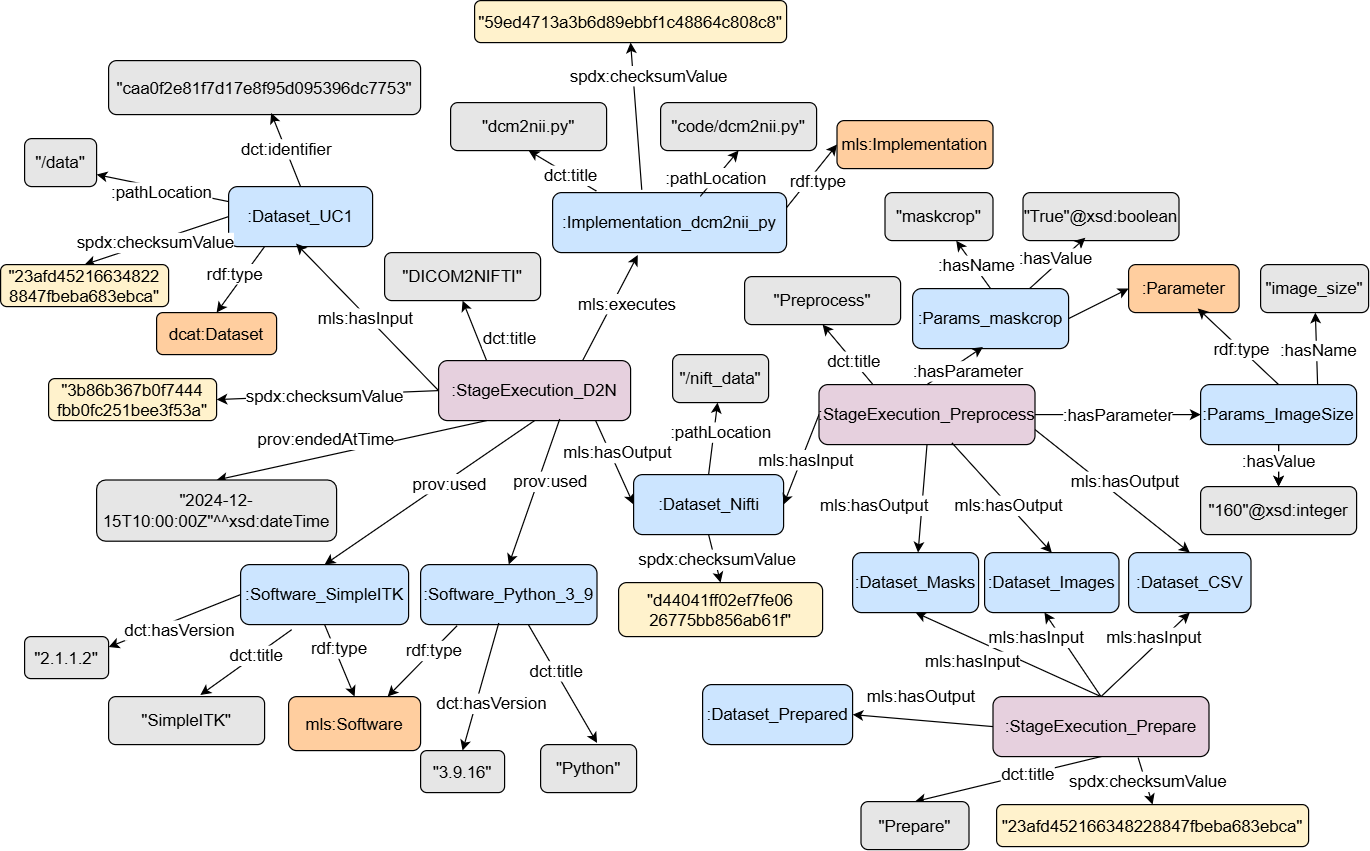}
  \caption{An example of the information logged on each AI preprocessing stage}
  \label{fig:preprocess_stage}
\end{figure}

Figure~\ref{fig:preprocess_stage} illustrates the provenance of the preprocessing pipeline used to prepare the input dataset for training. The process begins with the definition of a Study and an Experiment, which establish the context for a series of StageExecution activities. Each stage in the pipeline is clearly named and scoped—for example, DICOM2NIFTI, Preprocess, and Prepare—representing discrete data transformation steps. Input datasets (e.g., raw DICOM files) and expected outputs (e.g., NIfTI files, preprocessed masks and images, annotations) are explicitly linked to their corresponding execution stages using mls:hasInput and mls:hasOutput. For transparency and reproducibility, all scripts, datasets, and outputs are tracked with unique MD5 checksums (spdx:checksumValue), allowing version control and identification of the assets being generated. Each stage also references the software used (e.g., SimpleITK, Python) and its version along with input parameters influencing execution (e.g., image\_size, maskcrop).

\begin{figure}[!htb]
  \centering
  \includegraphics[width=1\textwidth]{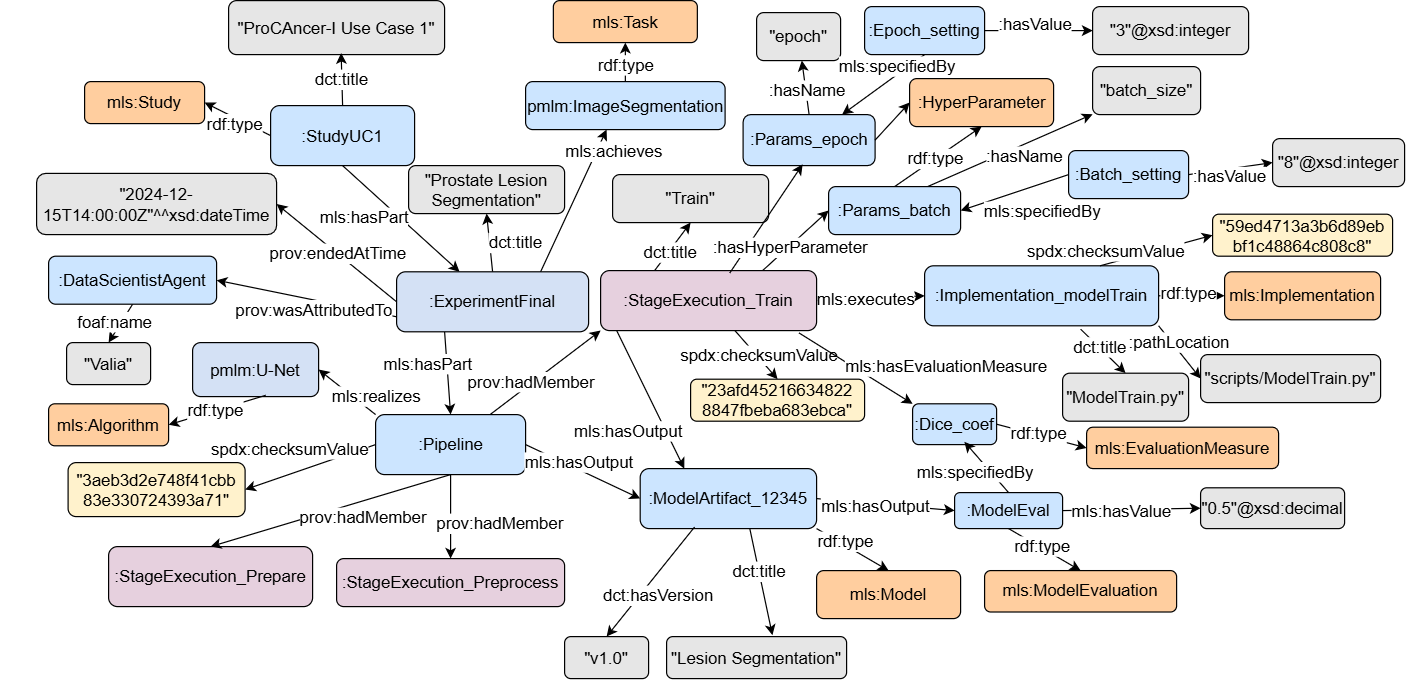}
  \caption{Instance-level provenance of a training workflow, showing hyperparameters used, evaluation measures, and the resulting model, compliant with PROV-O and MLS.}
  \label{fig:train_stage}
\end{figure}

\subsection{AI Model Training and Evaluation}

After the preprocessing steps are completed, the workflow advances to the model training phase. Similar to the previous stages, this process is captured using detailed provenance and metadata modeling. However, training introduces additional stage-specific metadata, including the definition and usage of hyperparameters (e.g., learning rate, batch size, number of epochs), the evaluation metrics employed to assess model performance (e.g., Dice coefficient), and the overall execution context such as code version and computational environment. These elements are crucial for reproducibility and traceability and are modeled using MLS constructs like mls:HyperParameterSetting, mls:EvaluationMeasure. The outcome of this stage is a trained model artifact, annotated with descriptive metadata such as title, version, and quality indicators. 


\subsection{AI Model in Production}

For registering the model in production, the developer must define a set of manually added metadata in a file automatically created by the AIPassport library, such as its intended purpose, potential threats (e.g., bias, misuse), license and ownership. These metadata are mandatory for the AI Model Passport, enabling developers and stakeholders to understand the model's use and intended purpose.

\subsection{AI Model Passport Marketplace}

\begin{figure}[h!]
  \centering
  \includegraphics[width=1\textwidth]{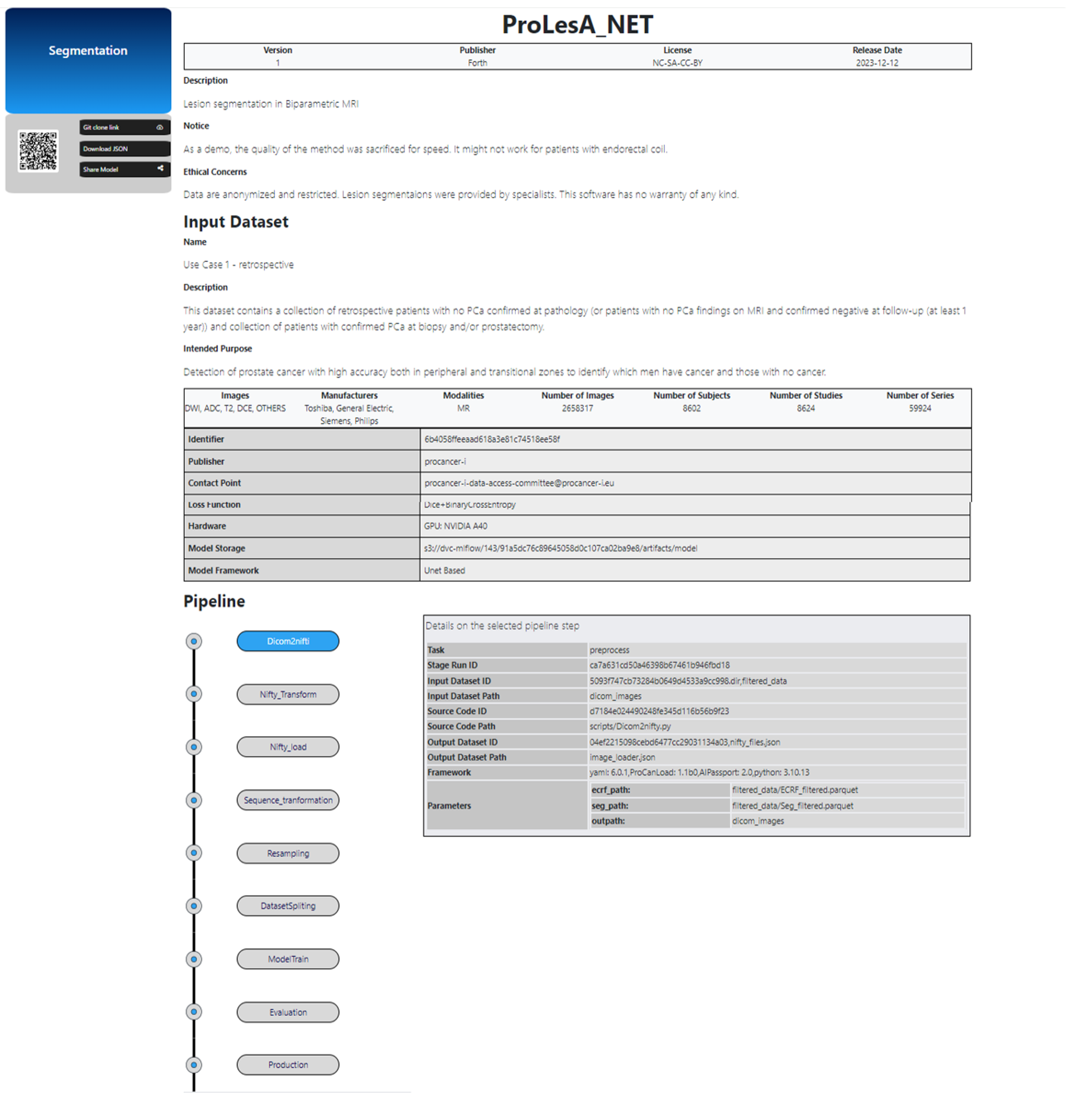}
  \caption{AI Model Passport marketplace.}
  \label{fig:marketplace}
\end{figure}

Apart from the machine-readable format of the AI Model Passport metadata, an AI model passport marketplace has been created in the context of the project as shown in Figure~\ref{fig:marketplace} , which is a web application responsible to create a human-readable version of the metadata from the registered AI models. Stakeholders and end-users may access the web page, to get all the provenance information about the registered models regarding information for the initial dataset, the AI system, and the development stages.  

Regarding the initial dataset, transparency is achieved by providing information regarding which use case was used, version, description, and intended purpose. It includes metadata related to the health domain, such as vendors, image modalities, sequence types, patients, and studies. It defines who the publisher of the dataset is and under which license and rights it is distributed. The model metadata consist of the AI method in use, hyper-parameters, optimiser, loss function, input and output size, the unique ID of the experiment, model, and the storage location. The DAG of the AI pipeline is visualized with all the stages. Each stage contains the data version (i.e., scripts, inputs, outputs), the used python packages, and the parameters. For the train and production stage, it also displays the model framework, metrics, and other information automatically logged by MLflow.




\section{Discusion}
In this work, we introduce the AI Model Passport framework, a standardized, machine-interpretable approach that addresses the growing demand for transparency, traceability, and accountability across the AI model development lifecycle. Our primary contribution lies in the integration of ontological metadata modeling with automated MLOps tooling to support the creation of verifiable digital identities for AI models and datasets. This enables not only traceable and reproducible workflows but also ensures regulatory compliance and fosters trust among stakeholders by providing a transparent record of the model’s origin, evolution, and usage context.

We validated our framework through the implementation of AIPassport, an open-source tool that operationalizes the AI Model Passport concept. This tool was deployed within the context of the ProCAncer-I project, a large-scale, EU-funded initiative focused on developing robust AI solutions for prostate cancer diagnosis and treatment using multiparametric MRI data. Specifically, we applied AIPassport to track the development of lesion and prostate gland segmentation models that address diagnosis and tumor characterization use cases, trained in imaging and clinical data from more than 14,300 patients. Throughout this process, AIPassport automatically captured a set of essential provenance metadata, recorded training and validation details, and maintained the traceability of associated data pipelines, demonstrating its practical applicability in a high-stakes clinical setting. 

Our results highlight that the AIPassport framework not only ensures transparency and verifiability but also simplifies documentation workflows and improves collaboration across teams by enabling shared, interpretable records of model development. Importantly, by embedding transparency and traceability into the core of AI model development, AIPassport addresses one of the key barriers to clinical adoption of AI: trust. Clinicians and regulators are more likely to adopt and rely on AI systems when the AI developments process is clearly documented and is auditable, the data lineage is clear, and the model behavior can be validated against its development history. In this regard, our approach contributes to building trust in AI by offering a verifiable, standardized mechanism to document and inspect how models are built, trained, and deployed. Furthermore, the clinical impact of increasing trust is substantial. Trustworthy AI systems are more likely to be integrated into clinical workflows, where the cost of error is high and accountability is paramount. The AIPassport framework provides healthcare stakeholders with confidence that models have been developed following rigorous traceability mechanisms with full documentation of the assumptions, limitations, and changes across their lifecycle. By enhancing traceability and auditability, we pave the way for improved quality control, post-deployment monitoring, and ultimately, safer and more effective AI-assisted clinical decision-making. 

\begin{table}[ht]
\centering
\caption{Comparison of AIPassport with existing AI transparency and traceability tools/frameworks.}
\label{tab:comparison}
\resizebox{\textwidth}{!}{%
\begin{tabular}{@{}lccccc@{}}
\toprule
\textbf{Framework/Tool} & \makecell{\textbf{Metadata} \\ \textbf{Automation}} & 
\makecell{\textbf{Lifecycle} \\ \textbf{Coverage}} & 
\makecell{\textbf{Model Identity} \\ \textbf{Verification}} & 
\makecell{\textbf{Regulatory} \\ \textbf{Alignment}} & 
\makecell{\textbf{Open} \\ \textbf{Source}} \\  \\ \midrule
Datasheets for Datasets & $\times$ & Partial & $\times$ & $\times$ & $\checkmark$  \\
Model Cards              & $\times$ & Partial & $\times$ & $\times$ & $\checkmark$  \\
MLflow                   & $\checkmark$ & Partial & $\times$ & $\times$ & $\checkmark$  \\
OpenLineage              & $\checkmark$ & Data Pipelines & $\times$ & $\times$ & $\checkmark$  \\
IBM FactSheets           & $\times$ & Partial & $\times$ & Partial & $\times$ \\
Azure ML / Vertex AI     & $\checkmark$ & Full & $\times$ & $\checkmark$ & $\times$ \\
ML Commons Croissant     & $\times$ & Datasets only & $\times$ & Partial & $\checkmark$ \\
\textbf{AIPassport} & $\checkmark$ & Full & $\checkmark$ & $\checkmark$ & $\checkmark$ \\ \bottomrule
\end{tabular}
}
\end{table}

When compared to existing systems and frameworks, AI Model Passport uniquely combines structured semantic modeling with end-to-end lifecycle coverage and digital identity verification. Table~\ref{tab:comparison} presents a comparative summary of our system alongside major existing tools and frameworks, focusing on key features such as metadata automation, lifecycle coverage, model identity verification, and regulatory alignment. As shown, AIPassport surpasses existing tools by offering a uniquely comprehensive and standards-based approach to transparency and traceability in AI development. In contrast to human-centered documentation frameworks such as Datasheets for Datasets and Model Cards, which require manual input and offer no automation, AIPassport delivers automated metadata collection across all phases of the AI lifecycle. While platforms like MLflow and OpenLineage provide partial coverage focused on experiment or pipeline tracking, AIPassport supports full end-to-end lifecycle coverage, including data collection, curation, model training, evaluation and deployment. Crucially, AIPassport is the only solution in the landscape that provides digital data and model identities, enabling data and model verifiability and reproducibility in all stages of the development pipeline to be confidently established across institutions and deployments. Additionally, it is explicitly designed to meet regulatory alignment requirements, embedding machine-interpretable, standards-based metadata that complies with initiatives such as the European Health Data Space and the Digital Product Passport. Finally, unlike proprietary solutions, AIPassport is fully open source, making it accessible and extensible for other domains. By addressing all critical dimensions of trustworthy AI—transparency, accountability, reproducibility, and compliance within a single, unified framework, AI Model Passport offers a novel solution that consolidates existing best practices, addresses key gaps left by existing tools and offers a complete solution to lifecycle data and model transparency. 


\textbf{Limitations.}
Despite its strengths, our approach has certain limitations. First, the current implementation of AIPassport is designed for centralized AI model development workflows and does not yet support federated learning or other distributed training paradigms, where data and models are managed across multiple decentralized nodes.  Another limitation lies in the current lack of integrated model monitoring capabilities. While the framework supports metadata and performance tracking during AI model development, it does not yet include dynamic monitoring in production environments. Although a monitoring framework to assess model behavior after production has been implemented in the ProCAncer-I project, these components have not yet been integrated into the AI Model Passport and are planned for future development to support a more comprehensive provenance framework. Finally, although AIPassport has been validated in the domain of prostate cancer imaging, broader application and evaluation in other clinical contexts and AI modalities are required to assess its generalizability and robustness.

\textbf{Future Directions.}
Future work will focus on enhancing the generalizability and domain applicability of the AI Model Passport framework. A key priority is to extend support for federated learning and decentralized AI development, where model training occurs across distributed nodes while maintaining privacy and data sovereignty. This requires adapting the current metadata tracking and identity verification mechanisms to operate effectively in such environments. In parallel, we are actively working on expanding the use of AIPassport across multiple critical domains, including robotics, education, media, transportation, healthcare, active ageing, and industrial AI systems. As part of this effort, we are incorporating standardized risk-related metadata, drawing on the framework and guidelines developed in the FAITH European project \cite{faith}. This extension aims to systematically capture information related to ethical, psychological, social, and technical risks that may emerge throughout the AI lifecycle, thereby broadening the trustworthiness profile of AI systems beyond traceability and reproducibility alone.

\section*{Acknowledgments}
This work is supported by the ProCancer-I project, funded by the European Union’s Horizon 2020 research and innovation program under grant agreement No 952159, and reflects only the authors’ view. The Commission is not responsible for any use that may be made of the information it contains.

\section*{Declaration of generative AI and AI-assisted technologies in the writing process}
During the preparation of this work the author(s) used ChatGPT (OpenAI) in order to improve clarity and flow. After using this tool/service, the author(s) reviewed and edited the content as needed and take(s) full responsibility for the content of the publication.

\bibliographystyle{unsrt} 
\bibliography{references} 
\clearpage

\end{document}